\documentclass[conference]{IEEEtran}
\IEEEoverridecommandlockouts
\usepackage{cite}
\usepackage[ruled,vlined]{algorithm2e}
\usepackage{amsmath,amssymb,amsfonts}
\usepackage{algorithmic}
\usepackage{graphicx}
\usepackage{multirow}
\usepackage{textcomp}
\usepackage{marvosym}
\usepackage{subcaption}
\usepackage{xcolor}
\usepackage{booktabs}
\usepackage{makecell}
\usepackage{xcolor, soul}
\usepackage{url}

\def\BibTeX{{\rm B\kern-.05em{\sc i\kern-.025em b}\kern-.08em
    T\kern-.1667em\lower.7ex\hbox{E}\kern-.125emX}}

\begin{document}

\title{FADACS: A Few-Shot Adversarial Domain Adaptation Architecture for Context-Aware Parking Availability Sensing}

\author{\IEEEauthorblockN{Wei Shao\textsuperscript{*}, Sichen Zhao\textsuperscript{*}\thanks{\textsuperscript{*}Equal contribution}, Zhen Zhang, Shiyu Wang, Mohammad Saiedur Rahaman, Andy Song, Flora D. Salim}
\IEEEauthorblockA{\textit{School of Computing Technologies} \\
\textit{RMIT University}\\
Melbourne, VIC, Australia \\
\text{[}firstname.lastname\text{]}@rmit.edu.au}
}

\maketitle
\begin{abstract}
Existing research on parking availability sensing mainly relies on extensive contextual and historical information. In practice, the availability of such information is a challenge as it requires continuous collection of sensory signals. In this study, we design an end-to-end transfer learning framework for parking availability sensing to predict parking occupancy in areas in which the parking data is insufficient to feed into data-hungry models. This framework overcomes two main challenges: 1) many real-world cases cannot provide enough data for most existing data-driven models, and 2) it is difficult to merge sensor data and heterogeneous contextual information due to the differing urban fabric and spatial characteristics. Our work adopts a widely-used concept, adversarial domain adaptation, to predict the parking occupancy in an area without abundant sensor data by leveraging data from other areas with similar features. In this paper, we utilise more than 35 million parking data records from sensors placed in two different cities, one a city centre and the other a coastal tourist town. We also utilise heterogeneous spatio-temporal contextual information from external resources,  including weather and points of interest. We quantify the strength of our proposed framework in different cases and compare it to the existing data-driven approaches. The results show that the proposed framework is comparable to existing state-of-the-art methods and also provide some valuable insights on parking availability prediction.
\end{abstract}


\section{Introduction}
Parking availability sensing plays a vital role in urban planning and city management~\cite{ranjan2019city}. According to a recent study, drivers spend more than 100,000 hours per year looking for parking slots in US \cite{shoup2017high}. Moreover, seeking available parking can lead to severe traffic congestion and air pollution \cite{kan2018estimating}. However, effective parking availability sensing can help drivers find a vacant parking spot. For example, drivers are likely to find a vacant parking slot if they know of parking availability ahead of time and choose a destination with low parking occupancy. This also helps governments to take appropriate measures based on understanding the utilisation of parking facilities, and provide more on-street parking lots in areas with high parking demand.

Parking dynamics have been studied in many research domains. In recent times, two types of sensing systems, explicit and implicit, have been used to infer parking availability in cities. The explicit sensing systems take a direct approach to  measuring parking occupancy through physical sensors such as underground sensors, RFID sensors, and monitoring cameras. In contrast, implicit sensing systems use an indirect approach to measure parking occupancy, e.g., through the sensing of contextual information such as weather, the number of restaurants and office buildings nearby, and density of public transportation stops \cite{yang2019deep}.

Most existing data-driven solutions rely heavily on long-term and historical data which is not always available in real-world scenarios \cite{qin2018eximius}. In recent times, several works have leveraged transfer learning techniques to estimate the traffic in areas without much historical data \cite{wang2018smart,ijcai2019-262}. However, domain shift and unsupervised learning remain as two main challenges in these studies. Another common challenge is that most existing works focus on temporal dependency of the contextual information and parking records. However, spatial dependency also plays a key role in parking occupancy, because the status of a parking slot is highly correlated with nearby parking slots. For example, drivers tend to park in a spacious space rather than a narrow space since small parking spaces are more likely to be difficult to park in. Also, low occupancy areas are preferred because of Nash equilibrium \cite{kokolaki2013efficiency}. Therefore, consideration of both spatial and temporal dependency is essential to parking occupancy prediction. 

Adding to the above challenges is the highly diverse feature space in the source and the target domains when the sensor data are collected from different cities. This paper, in particular, addresses a challenge that is often present when sensors in different cities are deployed by local authorities and the data are collected by different agencies, capturing local contextual information that is pertinent to the local urban fabrics with their specific characteristics. In this study, the source city is the city centre of an Australian state's capital with its Central Business Districts (CBD) areas, and the target city is a small coastal town mainly populated by retirees and is very popular with tourists. Hence, the parking patterns across the two regions are highly diverse and may not be directly transferable.

To overcome the challenges of integration of spatial dependency and temporal dependency and shared features extraction, we designed a domain adaptation architecture called FADACS, which can learn parking occupancy patterns without much historical parking data by utilising contextual sensor and parking sensor data from other areas. We adapted the idea from the computer vision area \cite{Tzeng_2017_CVPR} and incorporated with meta-sensing technologies. Specifically, we propose a generative adversarial convolutional-networks long short-term memory model for parking occupancy prediction by combining the generating ability of generative adversarial networks (GAN) and the spatio-temporal forecasting ability of convolutional long short-term memory (ConvLSTM). Compared to existing transfer learning models \cite{ijcai2019-262}, GAN-based transfer learning can easily learn the shared features of the source domain (where historical data is available and rich) and the target domain (where we would like to predict parking occupancy with no historical parking data) using the adversarial learning mechanism, and it does not need historical data from the target area. Additionally, the ConvLSTM model applies the convolution operations on the spatial domain and recurrent layers to the temporal domain \cite{ijcai2019-262}. We embed such a model into our adversarial learning framework and test it on two different real-world parking datasets with contextual information. The experimental results show that our proposed model is better than other existing transfer learning models for most cases. We also show that the contextual information has a significant influence on the prediction accuracy. In particular, the contributions of this paper are as follows:
\begin{itemize}
    \item To the best of our knowledge, we are the first to propose a GAN-based spatio-temporal transfer learning framework to predict parking occupancy in areas without historical parking records by utilising parking data from other areas and contextual information. We compare our proposed model with traditional transfer learning models which only take temporal information into consideration and state-of-the-art models such as ConvLSTM, which considers both spatial and temporal information but only uses parameter transferring approach to learn the distribution from the source domain. The experiments validate that our work, which incorporates both spatial information and temporal information and leverages the GAN-based transfer learning framework, outperforms other works in most scenarios. 
    \item We conduct an in-depth analysis of contextual factors that have potential influence on parking occupancy in different regions. We conducted the quantitative investigation on parking sensing by both implicit and explicit methods, and discovered insights on the contextual factors that have potentially influence parking occupancy.
\end{itemize}


\section{Related Work}
\textbf{Parking prediction:}
Many methods are appropriate in parking availability prediction, which can be treated as a time-series issues. Yu \textit{et al.}\cite{yu_real_2015} verified the effectiveness of real-time parking availability prediction using a time-series model. They established a variant of the autoregressive integrated moving average (ARIMA) model to predict remaining berths in an underground parking lot at Xinjiekou in Nanjing, China. Except the linear methods, Chu \textit{et al.} \cite{pengzi_service_2017} adopted the backpropagation neural network (BPNN) model proposed by Haviluddin \textit{et al.} \cite{haviluddin_daily_2014} on available parking spaces data collected in Xi'an, China. BPNN makes a nonlinear mapping between inputs and outputs, and the results show that it can generate effective predictions for parking lots with different capacities. Shao \textit{et al.} \cite{li_parking_2019} further utilised a large real-world parking spaces dataset from Melbourne, Australia and trained a long short-term memory (LSTM) model on that dataset.

\textbf{Domain adaptation:} Most of the machine learning models assume that the overall of training and test data are IID (independent and identical distributed), which is not always the case in the real world \cite{pan2009survey}. One major drawback is that test data which comes from a shifted distribution mostly leads to an unexpected performance drop. Except for the traditional approach, which is to build a new model and re-train that model, transfer learning is widely used to overcome this problem due to its better performance and efficiency. Transfer learning enables us to apply knowledge from the source domain upfront and to a new, related data or target domain. 

Pan and Yang \cite{pan2009survey} considered transductive transfer learning as similar to domain adaptation, which has the ability to transfer knowledge from a labelled source domain to an unlabelled target domain. Generative Adversarial Networks (GANs) \cite{goodfellow2014generative} since 2014 have become one of the hottest concepts in the field of artificial intelligence. Ganin \textit{et al.} \cite{ganin2016domain} first added an adversarial mechanism to domain adaptation and proposed a new framework called DANN (Domain-Adversarial Neural Network).  The objective of this method was to generate features that contribute to classification the most while making the discriminator unable to determine the source of the samples. In addition to mapping source and target domain into the same feature space, fine-tuning is another area of transfer learning, which is a process of reusing the training model in a second similar task and is also a simple method of transferring knowledge. Yosinski \textit{et al.} \cite{yosinski2014transferable} first demonstrated the transferability of features from a neural network. Since then, fine-tuning has been widely used in multiple areas. Google BERT \cite{devlin2018bert} is considered as a milestone in NLP, which comprises a pre-training stage and a fine-tuning stage. Similarly, to detect pathological brain in magnetic resonance images (MRI), the authors of \cite{lu2019pathological} achieved higher performance by using parameter-transfer learning based on a pre-trained AlexNet model. Besides improved performance and reduced training time, another advantage of fine-tuning is that it can counter the over-fitting problem that usually occurs on small datasets. Facing a 'cold-start' problem on expanding a market into a new city, Guo \textit{et al.} \cite{guo2018citytransfer} devised a new framework, called CityTransfer, which could learn inter-city and intra-city knowledge, based on collaborative filtering. One of the latest works applying few-shot learning technique to the area of sensing was proposed by Gong \textit{et al.} \cite{gong2019metasense} who used a transfer learning technique to learn user behaviours with only a few samples, although they did not apply the model to spatio-temporal data and contextual information.

In this paper, due to the lack of historical data on the target domain, we chose to approach the problem with domain adaptation, allowing the transfer of knowledge from the source domain to the target domain. 

\section{Data}
The two cities that were investigated in this research are the City of Melbourne and the town of Rye, both are in the state of Victoria, Australia. Melbourne is the capital city of Victoria. The municipality of Melbourne, with an estimated of 178,955 residents \cite{CBD_data}, has nearly 1 million people on average per day, visiting the municipality for work, education, travel or tourism \cite{shao2017traveling}. On the other hand, Rye is a small coastal town, part of the Mornington Peninsula Shire municipality. Rye had a population of approximately 8,416 in the 2016 census and is located about 100 km from the City of Melbourne. The Mornington Peninsula Shire hosts about 7.5 million visitors per year \cite{Rye_data}, about $50\%$ of those would visit Rye as one of their destinations, requiring parking spots, as driving is the main mode of transport to get into these coastal areas. Therefore, the major datasets in this paper include parking sensor data, Point of Interest (POI) data, weather data, and geographical data. All the datasets used in this system come from the following platforms: the City of Melbourne Open Data \cite{datamelbourne}, Time and Date AS \cite{timeanddate}, Google Maps API, and a proprietary Mornington Peninsula Shire data platform.

\subsection{Melbourne On-street Parking Data}
Parking relative data is from the City of Melbourne Open Data \cite{datamelbourne}. We use the following data sets:
\begin{itemize}
    \item On-street Car Parking Sensor Data 2017 
    \item On-street Parking Bays 
    \item On-street Parking Bay Sensors 
\end{itemize}

In this section, we will use the names Parking Sensor Data, Polygon Data and Location Data to represent each of the above datasets.

\subsubsection{Parking Sensor Data}
The Parking Sensor Data has 35.9 million records of 2017 on-street car parkings in Melbourne, containing 35 areas, 5044 sensor devices and 4695 parking slots. The numbers of the sensor devices and parking slots do not match up because if a sensor device needs to be removed by a fault, low battery or to upgrade the firmware, then a new sensor device with a different ID will replace the old one. In addition, Open Data indicated that these are streaming data. Therefore, regardless of whether the parking slot is occupied, each sensor runs for the whole day and continuously generates records. If the parking slot is occupied, the corresponding sensor will record the arrival time and departure time. Otherwise, during some periods, the sensor will automatically record the times and label it as non-occupied. Additionally, each midnight all sensors will upload their data and restart recording. The detailed format of the Parking Sensor Data is shown in Table \ref{tab:parking_data_column}.

\begin{table*}[htbp]
\begin{center}
\caption{Comparison of the Format for Both Parking Datasets}
\begin{tabular}{c||cc||l}
\hline\hline 
 \multirow{2}{*}{Column} & \multicolumn{2}{c||}{Dataset} & \multirow{2}{*}{Description} \\
  & Melbourne & Rye &   \\
\hline\hline 
 Device Id & ${\surd}$ & ${\surd}$  & The unique ID for the parking sensors \\
Arrival Time & ${\surd}$ & ${\surd}$ & \begin{tabular}[c]{@{}l} Date and time that sensor detected a vehicle\\ located over it\end{tabular} \\
Departure Time & ${\surd}$ & ${\surd}$ & \begin{tabular}[c]{@{}l} Date and time that sensor detected a vehicle no\\ longer located over it\end{tabular}   \\
Duration & ${\surd}$ & ${\surd}$ & \begin{tabular}[c]{@{}l} Time difference between arrival time and\\ departure time events, measured in seconds \end{tabular} \\
Overstay Duration & ${\times}$ & ${\surd}$ &  Time that a vehicle overstay, measured in seconds \\
In Violation & ${\surd}$ & ${\times}$ &  \begin{tabular}[c]{@{}l} Boolean value, indicate whether the parking event\\  is violation or not \end{tabular} \\
Street Name & ${\surd}$ & ${\surd}$ &  Name for the street of sector that a sensor locates  \\
Street Id & ${\surd}$ & ${\times}$ & A unique ID for streets \\
Street Marker & ${\surd}$ & ${\times}$ &  A unique ID for each parking slot  \\
Device Name & ${\times}$ & ${\surd}$ & Name for each device  \\
Sign/Restriction & ${\surd}$ & ${\surd}$ & \begin{tabular}[c]{@{}l} Parking rule/sign in effect at the time of the\\ parking event\end{tabular} \\
Longitude & ${\surd}$ (via Location Data) & ${\surd}$ &  The longitude of the parking sensor \\
Latitude & ${\surd}$ (via Location Data) & ${\surd}$ &  The latitude of the parking sensor \\
 \hline\hline
\end{tabular}
\label{tab:parking_data_column}
\end{center}
\end{table*}


\subsubsection{Polygon Data}
There are 24,074 records in this dataset, including all parking slots in the Melbourne area. Each record contains a series of locations that define the actual boundary of a parking slot. Although each polygon has its unique parking bay ID, only a small portion of them have a built-in sensor with a street marker ID that can link to the parking data mentioned above.

\subsubsection{Location Data}
Since the polygon data contains the boundary of all parking slots regardless of whether they have dedicated sensor. In this paper, we also used another data called Location Data that contains a single longitude-latitude tuple for each parking slot which can eliminate ambiguous distance calculation. According to the recommendation from the City of Melbourne Open Data Platform \cite{datamelbourne}, Location data and Polygon data should be supplemented with Street Marker ID.

\subsection{Rye Data}
\begin{figure*}[t]
    \centering
    \includegraphics[width=0.9\textwidth]{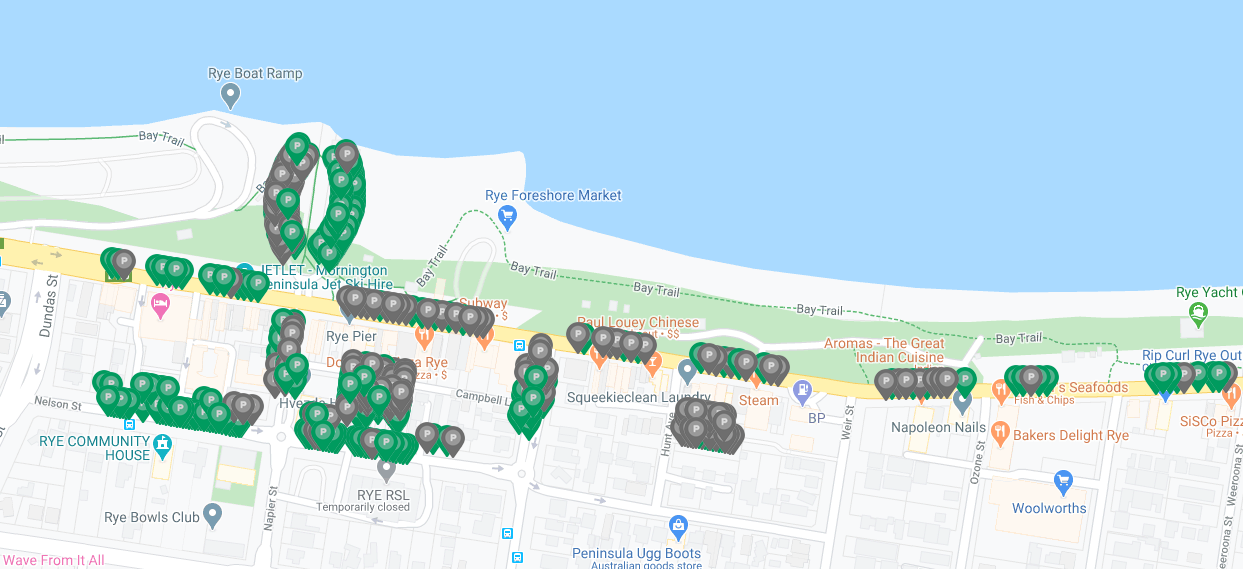}
    \caption{The location and status example of parking slots located in Rye. Green indicates available slots while grey stands for slots that are currently occupied.}
    \label{fig:rye_map}
\end{figure*}

This data was collected by the Mornington Peninsula Shire, which includes 
179,288 records across 527 devices in Rye, Victoria. 
The time range of this data is from 17th Nov 2019 to 20th Feb 2020 and spatially spread over 7 sectors. Details can also be found in Table \ref{tab:parking_data_column}, and an example of the status for those parking slots is shown in Fig \ref{fig:rye_map}.

\subsection{Point of Interest Data}
The Point of Interest (POI) data of Melbourne came from the City of Melbourne's Open Data Platform \cite{datamelbourne} under project CLUE (Census of Land Use and Employment). It records comprehensive information about land use and is updated frequently. We chose three sub-datasets that cover most of the possible POI categories that relate to parking prediction: 

\begin{itemize}
    \item Bars and pubs, with patron capacity \cite{datamelbournebar} 
    \item Cafes and restaurants, with seating capacity \cite{datamelbournecafe} 
    \item Landmarks and places of interest, including schools, theatres, health services, sports facilities, places of worship, galleries and museums \cite{datamelbourneplandmark}
\end{itemize}

\subsubsection{Bar and Cafe Data}
The first two datasets record all business establishments for pubs, bars, cafes and restaurants. The data collection of this part starts in 2002 and is updated annually. We combined them due to their similar structure. After filtering out the data in 2017, we got 263 and 3563 records, respectively. Each record contains the trading name for that business establishment, a street address and a coordinate which can be pinned on a map.

\subsubsection{Landmarks and places of interest}
The structure of the last dataset, landmarks and places of interest, is different from the other two, and only has 242 records with coordinate information and theme. There are 49 themes, such as hostel, cinema, library and casino.

\subsection{Weather Data}
Weather data from the two locations were collected from Time and Date AS \cite{timeanddate}. We gathered the weather data for Melbourne and Mornington according to the time range of the data we gathered the two locations. Detailed columns used in this paper are shown below: 

\begin{itemize}
    \item Time: The specific time with the weather information
    \item Temp: The temperature in Celsius scale
    \item Wind: The wind speed measured in km per hour
    \item Barometer: The barometer in millibars
\end{itemize}

\section{Data Pre-processing}
\subsection{Matching the Parking Slot with Location}
The first step in pre-processing is to match each parking slot with the correct location coordinate. We decided to use the Street Marker ID as the primary key for matching since the Device ID of a parking slot could be changed. For parking slots in the Melbourne area, we joined the three aforementioned files on the StreetMarker column. If the location coordinate was missing for a parking slot, it was filled with the centre of its polygon. To ensure the validity of this method, we checked to ensure that all existing location values fall into their corresponding polygon boundaries. For parking slots in the Mornington Peninsula, there was no need for this step since all slots have a corresponding location coordinate.

\subsection{Parking Slot Area Grouping}
In this paper, instead of the individual parking slot, we used parking lots as the base unit in our experiments. A parking lot is defined as a cluster of parking slots that fall into the same locality and share the same parking restriction rules \cite{shao2016clustering, shao2020incorporating}. The size of a parking lot is relatively small. We used this particular setup because the data collected directly from the sensor is too noisy. Cleaning the data and grouping them together while keeping a relatively small lot size, allow us to mitigate the possible interference coming from the raw data. More details will be discussed in Section \ref{anomolies}. We switch the objective of the model from predicting whether a given parking slot is occupied or not at a certain time step to the occupancy rate of a group of parking slots. Since the status of a single parking slot is noisy, this approach simplifies the problem while still maintaining the original goal of parking availability sensing.

\subsubsection{Melbourne Data}
\begin{figure*}[!tbp]
\centering
\begin{subfigure}[b]{0.45\textwidth}
    \includegraphics[width=\textwidth]{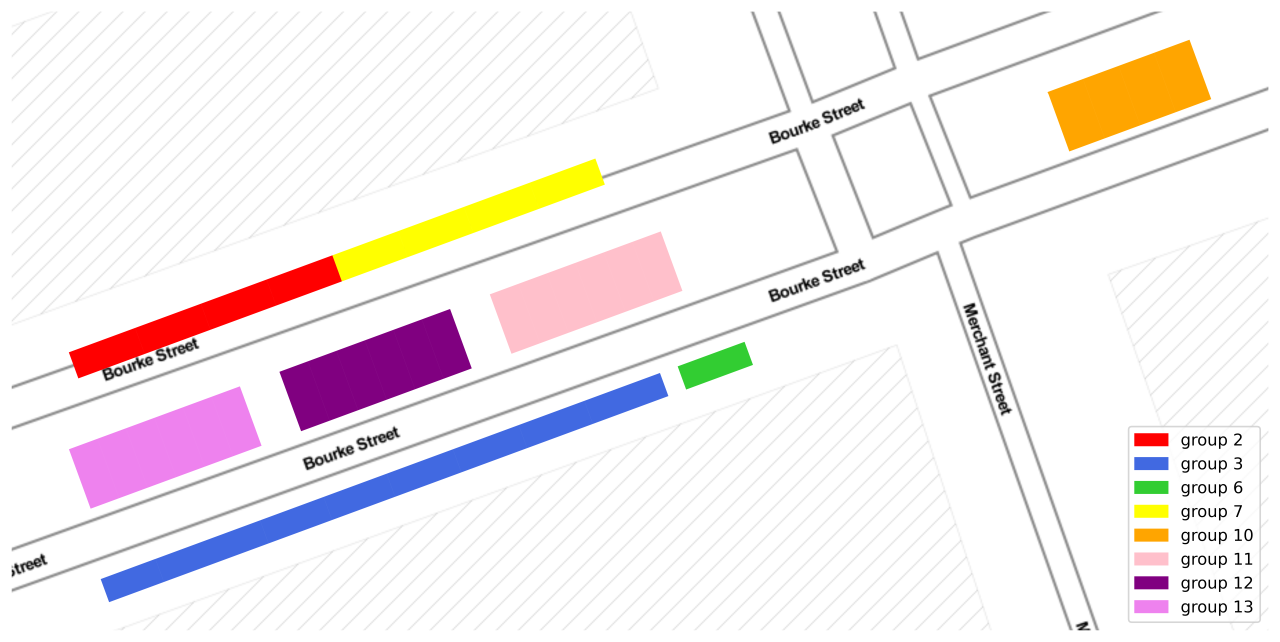}
    \caption{Clusters connection only. This is one small sample of parking slots in the Docklands area. Each rectangle represents a parking slot. Group 2 (red) and Group 7 (yellow) are close to each other and have the same parking rule, but are not in the same group.}
    \label{fig:cluster_polygon}
  \end{subfigure}
  \hfill
  \begin{subfigure}[b]{0.45\textwidth}
    \includegraphics[width=\textwidth]{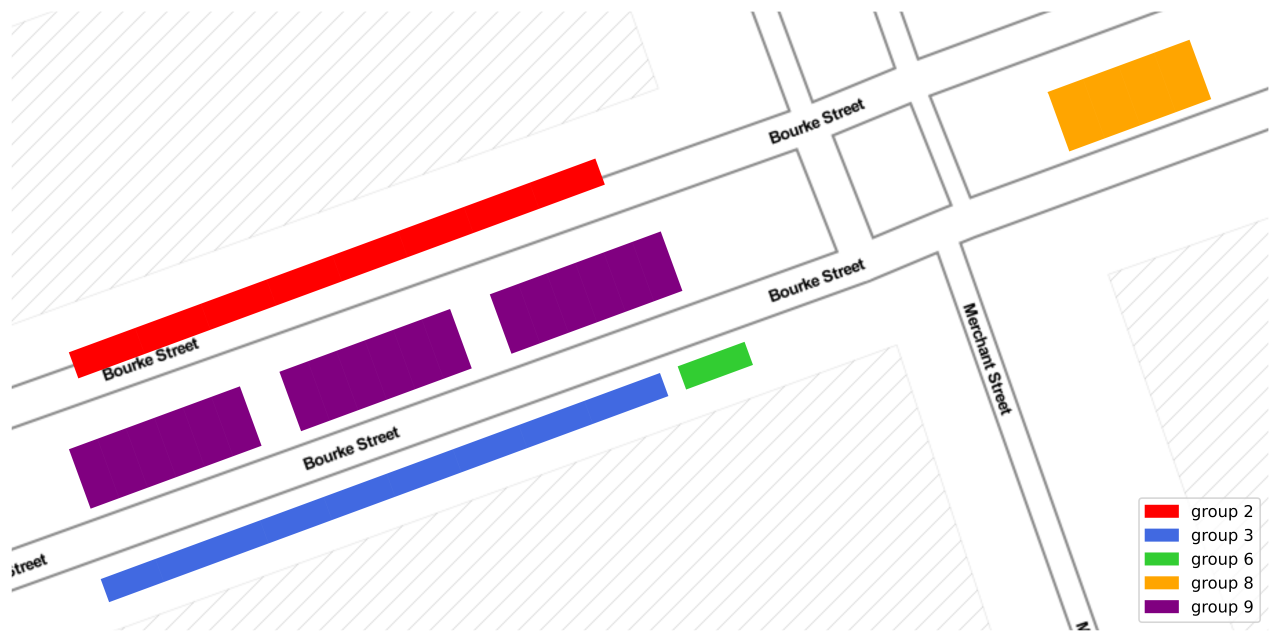}
    \caption{Clusters distance and rules. If the distance of two parking slots is within the given threshold and they have the same parking rules, they are in the same group. The original Group 2 and Group 7 are now in the same group (red), and group 11, 12 and 13 are in the same group (purple).}
    \label{fig:cluster_threshold}
  \end{subfigure}
  \hfill
  \begin{subfigure}[b]{0.6\textwidth}
    
        \includegraphics[width=\textwidth]{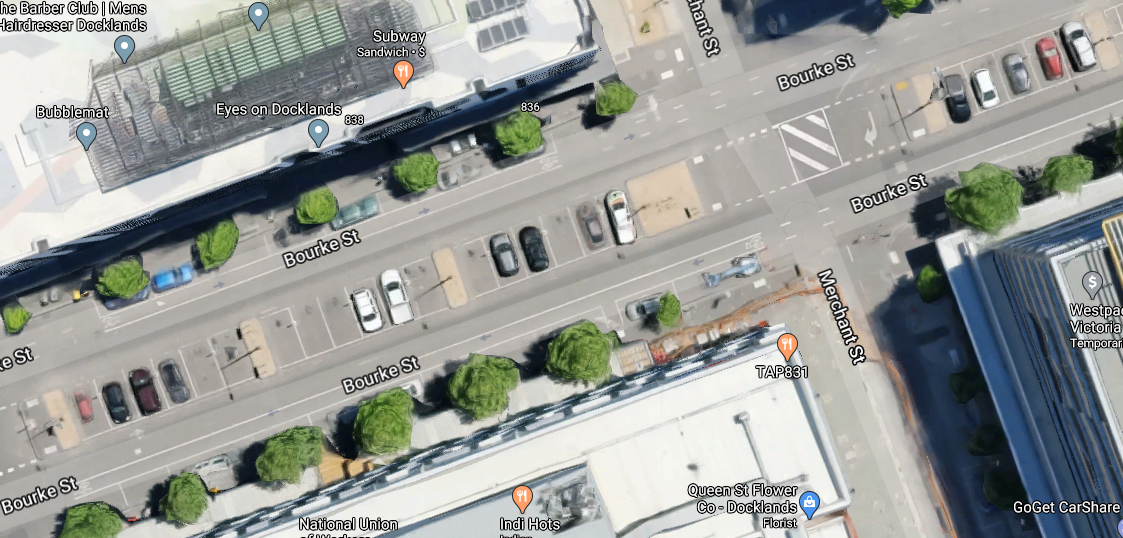}
    \caption{The same area in Google Satellite Map}
    \label{fig:google_satellite}
    
  \end{subfigure}
  \caption{A Sample of Parking Slot Clusters in Melbourne CBD}
\end{figure*}

Since each parking lot was considered the base unit for later stages, we needed to ensure that a consistent parking rule is shared within each lot.
Due to reasons such as construction, the rules could change during the study period. The parking rules for October, November and December 2017 were replaced with whole-year rules. We then create an initial grouping which groups those spatially connected slots according to their polygon boundaries. 
However, this initial grouping result which used only the geometric information still needed improvement. As shown in Fig. \ref{fig:cluster_polygon}, some of the parking slots are not under the same parking lot, even though they are close to each other and have the same parking restriction. Based on this finding, we performed the grouping operation based on three criteria: connection, distance and rules. Spatially connected slots were clustered into the same lot. For those that are not connected, if they had the same parking restriction and the distance between them was under a threshold, they were also put into the same lot. To calculate this threshold, we selected a specific area and set the value as the sum of the mean connection distance and 1.5 times its standard deviation. The example grouping result is shown in Fig. \ref{fig:cluster_threshold}, and we clustered all 4192 parking slots in the Melbourne data into 912 separate parking lots.

\subsubsection{Rye Data}
It was much simpler to cluster parking slots for the Rye dataset. Although there is no polygon information in that dataset, each parking slot in the Rye parking dataset has a coordinate with consistent parking restriction information. We first grouped the data by sector and rule information to reduce complexity. For each group in the same sector with the same rule, we checked the distance between two neighbour groups. If the distance was smaller than the threshold that we use in the Melbourne dataset, we combined them to get a larger group. 

\subsection{Occupancy Rate Calculation}
\label{anomolies}

In order to extract the occupancy of a parking lot at a given time, we removed irrelevant records showing no vehicle present or belonging to one of the following anomalies:
\begin{itemize}
    \item DurationSeconds is non-positive, which is usually caused by a faulty sensor;
    \item ArrivalTime and DepartureTime are both at exactly midnight;
    \item DepartureTime is past the midnight of the ArrivalTime;
    \item Records overlapping with other records, which could be caused by other unexpected interference.
\end{itemize}
This is a fairly important cleaning process step, since over half of the parking data was corrupted under these definitions and needed to be eliminated. Then, we sliced the data every 1 and 5 minutes and calculated the occupancy of each parking lot at that time. The occupancy rate $Oc_l^t$ is represented by a real value from 0 to 1 which indicates the proportion of slot was occupied in any given lot $l\in \{L\}$ at timestamp $t$. For example, if there is a parking lot $l$ that contains ten slots and at timestamp $t$, 5 of them are occupied, the occupancy rate $Oc_l^t=5/10=0.5$.

\subsection{Contextual Features}

Based on the POI and Weather dataset that we collected, we calculate a series of contextual features shown in Table \ref{tab:contextual_features}.

\begin{table}[]
    \centering
        \caption{Detailed Description of Contextual Features.}
  \begin{tabular}{c||ll}
        \hline\hline 
        Feature Name & Description & Feature Type \\
        \hline\hline 
        \begin{tabular}[c]{@{}c}Num of open\\poi 1.0\end{tabular}  & \begin{tabular}[c]{@{}l}Number of Open POIs within \\ 1.0 km nearby\end{tabular}  & POI\\
        \begin{tabular}[c]{@{}c}Num of open\\poi 0.5\end{tabular} & \begin{tabular}[c]{@{}l}Number of Open POIs within \\ 0.5 km nearby\end{tabular} & POI\\
        \begin{tabular}[c]{@{}c}Num of poi \\1.0\end{tabular}        &  \begin{tabular}[c]{@{}l}Number of POIs within 1.0 km\\ nearby\end{tabular} & POI\\
        \begin{tabular}[c]{@{}c}Num of poi \\0.5\end{tabular}      &  \begin{tabular}[c]{@{}l}Number of POIs within 0.5 km\\ nearby\end{tabular} & POI\\
        Min dis 1.0            & \begin{tabular}[c]{@{}l} The shortest distance of POIs nearby\\ within 1.0 km \end{tabular}  & POI\\
        Min dis 0.5           &  \begin{tabular}[c]{@{}l} The shortest distance of POIs nearby\\ within 0.5 km \end{tabular} & POI\\
        Day of week             &  \begin{tabular}[c]{@{}l} The ordinal of the day in the\\  whole week\end{tabular} & date-time\\
        Day of month            & \begin{tabular}[c]{@{}l}The ordinal of the month in the\\ whole year  \end{tabular} & date-time\\
        Hour                  &  The hour of the day & date-time\\
        \begin{tabular}[c]{@{}c}Parking\\availability\end{tabular}          &  \begin{tabular}[c]{@{}l}Whether the parking lot is\\ currently available\end{tabular} & availability\\
        Temperature           &  The temperature in degree Celsius  & Weather\\
        Wind                  &  The wind speed & Weather\\
        Barometer             & The barometer value & Weather\\
        Humidity              &  The humidity value & Weather\\
        \hline\hline
    \end{tabular}
    \label{tab:contextual_features}
\end{table}

For the Point Of Interest features, we consider that the total number of POIs and the number of opening POIs within a given distance may have a higher impact on the occupancy of parking lots, since if there are many restaurants around a parking lot, this lot should be more popular during mealtime and have a lower occupancy rate at other periods.

There are a total of 4068 POIs in Melbourne after aggregating all three aforementioned datasets. To crawl the opening hours for all these places, we used two Google Maps APIs: Place Search and Place Details. The former provides a place\_id for each place which is used for searching the details in the latter. We crawl the opening hours for POIs in both Melbourne and Rye, and we got results for a total of 50 POIs within eight different sub-categories.

For a given parking lot at a specific date-time, we first calculated the distance between all POIs and this parking lot, then we extract the features based on the opening info retried in the former stage. We also record its minimum distance to an opening or any POI.

After the extraction, we applied an ANOVA (Analysis of variance) test on both datasets. The ANOVA test \cite{fisher1992statistical} is widely used as a statistical hypothesis testing methods. The result is statistically significant when the null hypothesis is rejected (the p-value is lower than a pre-set threshold). As shown in Tables \ref{tab:pcc_melb} and \ref{tab:pcc_rye}, the Pearson Correlation Coefficient is higher when there is a statistically significant correlation. Surprisingly, some contextual features have a opposite correlation with prediction occupancy in Melbourne CBD dataset and Rye dataset. For example, humidity has a negative correlation with parking occupancy in Melbourne city but has a positive correlation in the Rye area. This reflects the possible shift in the data distributions of those two datasets, hence proving the need to introduce a domain adaptation method in this situation.

\begin{table}[!ht]
\caption{ANOVA Test Results of Data from Melbourne City in Feb 2017}
\begin{tabular}{c||llll}
\hline \hline
Features& \begin{tabular}[c]{@{}l@{}}Pearson\\Correlation\\Coefficient\end{tabular} & F Value  & p-Value & Type  \\ 
\hline \hline
\begin{tabular}[c]{@{}c}Num of open\\poi 1.0\end{tabular} & 0.34& 51742.63 & 0 & POI      \\
\begin{tabular}[c]{@{}c}Num of open\\poi 0.5\end{tabular} & 0.33& 48513.95 & 0  & POI       \\
\begin{tabular}[c]{@{}c}Num of poi \\1.0\end{tabular}      & 0.05& 1149.31  & 1.48e-251 & POI\\
\begin{tabular}[c]{@{}c}Num of poi \\0.5\end{tabular}        & 0.09& 3356.62  & 0  & POI      \\
Min dis 1.0           & -0.04& 570.99   & 4.22e-126 & POI\\
Min dis 0.5           & -0.04& 570.99   & 4.22e-126 & POI\\
Day of week             & 0.07& 2063.57  & 0 & date-time\\
Day of month            & 0.01& 78.61    & 7.60e-19  & date-time\\
Hour                  & 0.16& 9896.81  & 0         & date-time\\
\begin{tabular}[c]{@{}c}Parking\\availability\end{tabular}          & 0.02& 108.92   & 1.70e-25  & availability\\
Temperature                  & 0.23& 21622.33 & 0         & Weather\\
Wind                  & 0.13& 6871.60  & 0         & Weather\\
Barometer             & -0.01& 22.81    & 1.79e-06 & Weather\\
Humidity              & -0.25& 26409.49 & 0        & Weather \\
\hline \hline
\end{tabular}
\label{tab:pcc_melb}
\end{table}

\begin{table}[]
\caption{ANOVA test results of data from Rye in Feb 2020}
\begin{tabular}{c||llll}
\hline \hline
Features& \begin{tabular}[c]{@{}l@{}}Pearson\\Correlation\\Coefficient\end{tabular} & F Value  & p-Value   \\ 
\hline \hline
\begin{tabular}[c]{@{}c}Num of open\\poi 1.0\end{tabular} & -0.32 & 27946.25 & 0 & POI         \\
\begin{tabular}[c]{@{}c}Num of open\\poi 0.5\end{tabular} & -0.30 & 24221.17 & 0 & POI         \\
\begin{tabular}[c]{@{}c}Num of poi \\1.0\end{tabular}       & 0.00  & 5.66     & 1.74e-02 & POI   \\
\begin{tabular}[c]{@{}c}Num of poi \\0.5\end{tabular}       & -0.03 & 245.34   & 2.86e & POI    \\ 
Min dis 1.0          & -0.12 & 3487.51  & 0 & POI         \\
Min dis 0.5          & -0.12 & 3487.51  & 0  & POI        \\
Day of week              & -0.06 & 771.68   & 1.41e-169 & date-time\\
Day of month             & 0.11  & 3032.63  & 0 & date-time        \\
Hour                  & -0.07 & 1139.34  & 3.41e-249 & date-time\\
\begin{tabular}[c]{@{}c}Parking\\availability\end{tabular}          & 0.02  & 118.98   & 1.07e-27  & availability\\
Temperature                  & -0.24 & 14862.64 & 0 & Weather        \\
Wind                  & -0.07 & 1165.69  & 6.82e-255 & Weather\\
Barometer             & 0.06  & 913.98   & 2.08e-200 & Weather\\
Humidity              & 0.19  & 9247.09  & 0 & Weather        \\

\hline \hline
\end{tabular}
\label{tab:pcc_rye}
\end{table}

\section{FADACS Architecture}

\begin{figure*}[t]
    \centering
    \includegraphics[width=\textwidth]{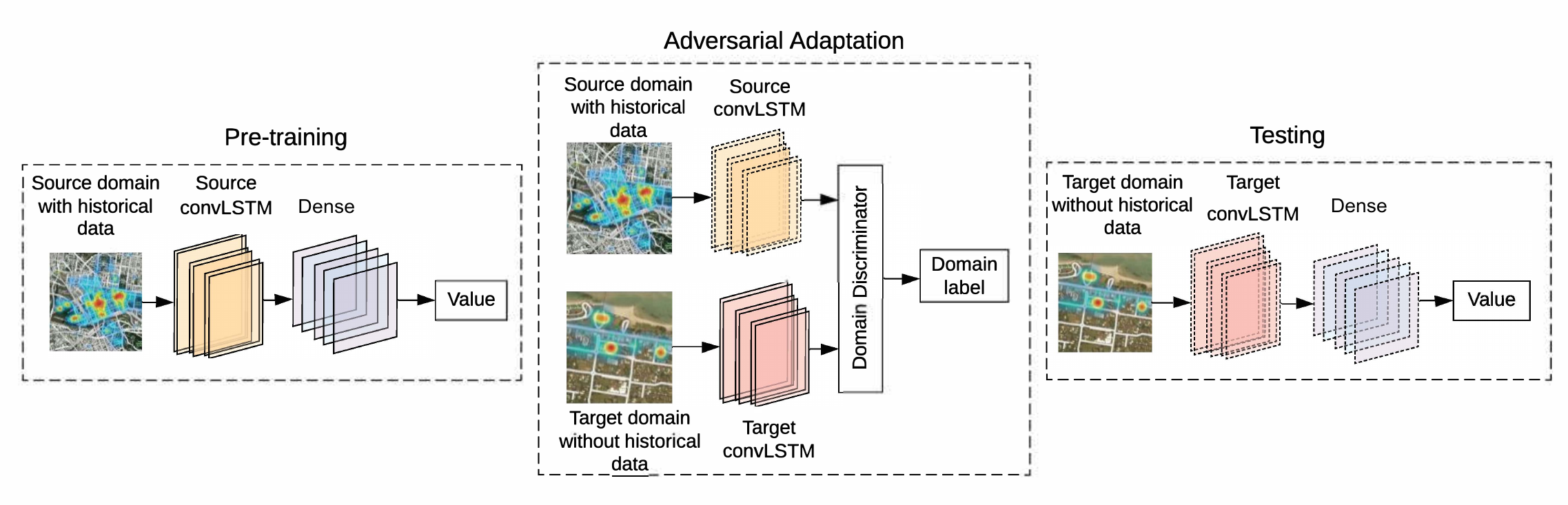}
    \caption{FADACS Domain Adaptation Architecture}
    \label{fig:transfer}
\end{figure*}

The traditional transfer learning method uses a kind of fine-tuning, which first loads a pre-trained parameter from other tasks and then re-trains them on the new domain/task. However, one issue that needs to be faced in real-world usage is that most tasks have little or no historical data. According to \cite{ganin2016domain}, Tzeng \textit{et al.} \cite{Tzeng_2017_CVPR} proposed a general architecture for adversarial domain adaptation named Adversarial Discriminative Domain Adaption (ADDA). Their framework using in ADDA combines a discriminative model, untied weight sharing and GAN loss together, which shows promising performance on unsupervised transfer learning. Unlike other domain adaptation methods, ADDA introduces an adversarial mechanism that trains an encoder to translate the features from the target domain to the latent space shared by both the source and target domains. Meanwhile, a discriminator is trained simultaneously to distinguish the origin of each latent code. 

In this study, we adopt the original ADDA framework, which is initially used for image classification tasks and modify it to makes it applicable to our time-series prediction problem. We use $X^{s}$ and $X^{t}$ to denote source and target domain features. $Y^{s}$ denotes occupancy rate of parking lots from the source domain. $M^{s}(X^{s})$ denotes source mapping/encoder and $M^{t}(X^{t})$ is the target mapping. The regression model is represented as $F$ while $D$ stands for the discriminator. The architecture we use in this paper is shown in Fig. \ref{fig:transfer}, and it comprises the three following stages. The first part is the pre-training step for the a source encoder $M^{s}(X^{s})$ and a regression model based on the source domain data. Similar to an auto-encoder structure, the encoder here learns a mapping of the source domain to a latent space. The regressor learns to decode features from this latent space and make a prediction on top of that. We used ConvLSTM (Convolutional Long-Short Term Memory) proposed in \cite{xingjian2015convolutional} as the encoder, which shows good performance on spatio-temporal data. Extending on a common LSTM unit, matrix multiplication is replaced by a convolution operation at each gate in the LSTM cell. The ConvLSTM calculations are shown in equation \ref{eq1} below: 

\begin{equation} \label{eq1}
\begin{split}
i_{t} & = \sigma (W_{xi} \ast X_{t} + W_{hi} \ast H_{t-1} + W_{ci} \circ C_{t-1} + b_{i})\\
f_{t} & = \sigma(W_{xf} \ast X_{t} + W_{hf} \ast H_{t-1} + W_{cf} \circ C_{t-1} + b_{f}) \\
C_{t} & = f_{t} \circ C_{t-1} + i_{t} \circ \tanh(W_{xc} \ast X_{t} + W_{hc} \ast H_{t-1} + b_{c}) \\
o_{t} & = \sigma(W_{xo} \ast X_{t} + W_{ho} \ast H_{t-1} + W_{co} \circ C_{t} + b_{o}) \\
H_{t} & = o_{t} \circ \tanh(C_{t})
\end{split}
\end{equation}

where '$\ast$' denotes the convolution operator and '$\circ$' denotes the Hadamard product. Besides, we use $C_1, ..., C_t$ and $H_1, ..., H_t$ to represent the cell output and hidden state for the given time step $t$ and $i_t, f_t, o_t$ different gates \cite{xingjian2015convolutional}.

The next step is an adversarial adaptation, which is to teach a target encoder $M^{t}(X^{t})$ so that the discriminator $D$ cannot distinguish the origin of the sample. By fixing the source encoder parameter, the adversarial loss is used to minimise the distance of the mapping between the source and target domain: $M^{s}(X^{s})$ and $M^{t}(X^{t})$ and maximise the discriminator loss. The loss function is as follows: 

\begin{equation}
\begin{split}
\min_{D} \mathcal{L}_{\text{adv}_{D}}(X^{s}, X^{t}, M^{s}, M^{t}) &= \\
    & - \mathbb{E}_{x_{s} \thicksim X_{s}}[\log D(M^{s}(X^{s}))] \\
    &    - \mathbb{E}_{x_{t} \thicksim X_{t}}[\log(1 - D(M^{t}(X^{t})))] \\
\min_{M^{t}} \mathcal{L}_{\text{adv}_{M}}(X^{s}, X^{t}, D) &= \\
    & - \mathbb{E}_{x_{t} \thicksim X_{t}}[\log D(M^{t}(X^{t}))]
\end{split}
\end{equation}

In the final stage, we assembled the learned target encoder $M^{t}(X^{t})$ and regression model $F$ together, and use data from the target domain to test its performance. The regressor should have the ability to generate quality prediction since the latent features from the target domain overlap with those from the source domain after the previous adaptation stage\footnote{The code is available at \url{ https://github.com/cruiseresearchgroup/FADACS_Parking_Prediction}}.

\section{Experiments}





\subsection{Experimental Settings}

\begin{table}[ht]
\caption{Summary of Architecture}
\label{tab:acrhitecture}
\begin{center}
\begin{tabular}{c||cc}
\hline\hline 
  & Layer & Parameters \\
\midrule\midrule 
 \multirow{4}{*}{\makecell[c]{Feature Extractor\\(For both source \\and target domain)}} & \multirow{4}{*}{ConvLSTM} & input: 16 channels \\
  &   & hidden: 200 channels \\
  &   & output: 60 channels \\
  &   & Sequence length: 6 \\
\hline 
 \multirow{2}{*}{Regressor} & Dense & (60, 1) \\
  & Sigmoid &  \\
\hline 
 \multirow{6}{*}{Discriminator} & Dense & (60, 100) \\
  & ReLU &  \\
  & Dense & (100, 100) \\
  & ReLU &  \\
  & Dense & (100, 1) \\
  & LogSoftmax &  \\
 \hline\hline
\end{tabular}
\end{center}
\end{table}

We conducted all of our experiments on a Linux Server (CPU: Intel Xeon Gold 6132 CPU @ 2.60GHz - 56 cores, GPU: NVIDIA Quadro V100). 
In order to find the best parameters, we used a parallel grid search strategy that utilises all cores in this Linux cluster. As stated in the Data Pre-processing section, we used 5 minutes as the basic interval between records. Additionally, each sample contained features from the recent 30 minutes (i.e., 6 data points for each sample), and the tasks are to predict the parking occupancy rate for the next 5, 15 and 30 minutes (the next 1, 3, 6 timesteps).  Two parking sensor datasets collected from Melbourne, Victoria and Rye, Victoria were used. The first covers a whole year time period (2017), while the second has a time range from 17th Nov 2019 to 20th Feb 2020. This reflects the big difference in both spatial and temporal domains which makes it difficult to apply the transfer learning method.

\subsubsection{Evaluation Metric}
Mean Absolute Errors (MAE) and Root Mean Squared Errors (RMSE) were used to evaluate the effectiveness of different models. Except for the adversarial adaptation stage, all models were trained using RMSE as the loss function. The calculation of MAE and RMSE are shown below: 

\begin{equation}
MAE=\frac{1}{TL}\sum ^{T}_{t=1}\sum ^{L}_{l=1}\Bigl|\left( Oc^{t}_{l} -\widehat{Oc^{t}_{l}}\right)\Bigl|
\label{equation_mae}
\end{equation}

\begin{equation}
RMSE=\sqrt{\frac{1}{TL}\sum ^{T}_{t=1}\sum ^{L}_{l=1}\left( Oc^{t}_{l} -\widehat{Oc^{t}_{l}}\right)^{2}}
\label{equation_rmse}
\end{equation}

where $Oc^{t}_{l}$ represents the parking occupancy rate for a given lot $l$ at timestamp $t=\{t=5, 15, 30 \  \text{mins}\}$.

\subsubsection {Baseline Models}
For FADACS, we implemented two variants:
\begin{itemize}
    \item ADDA (MLP): using MLP (Multi-Layer Perceptron) as the encoder to learn the mapping from the source/target domain to the latent space.
    \item ADDA (ConvLSTM): using ConvLSTM as the encoder to learn the mapping from the source/target domain to the latent space. The intention here is to extract better latent features using ConvLSTM since the problem here is a spatial-temporal prediction problem.
\end{itemize}

We compare FADACS with the following baselines:
\begin{itemize}
    \item HA (Historical Average): using the mean historical data as the prediction of the future data.
    \item MLP: (Multilayer Perceptron): a feed-forward neural networks which is widely used in function approximation and general regression problems. It also relies on feature extraction and is data hungry. It cannot distinguish between temporal features and spatial features.
    \item LSTM (Long Short-Term Memory): a recurrent-based method that is widely used in many time-series prediction tasks \cite{li_parking_2019}. But it only focuses on the temporal domain. Therefore, if the spatial domain also plays an important role, its performance will be limited. 
    \item ConvLSTM: a state-of-the-art methods used in transfer learning area that can utilise features from both spatial and temporal domains \cite{ijcai2019-262}. 
\end{itemize}

\begin{table}[t]
\caption{Performance comparison with full parking data before domain adaptation}
\label{tab:source}
\begin{tabular}{l||cc}
\hline \hline
Model                                & MAE (5/15/30 mins)       & RMSE (5/15/30 mins)      \\ 
\hline \hline
HA        & 0.0600                  & 0.1219                   \\
MLP       & 0.0536 / 0.0895 / 0.1188        & 0.0988 / 0.1456 / 0.1771 \\
LSTM      & 0.0419 / 0.0767 / 0.1011 & 0.0942 / 0.1443 / 0.1765 \\
ConvLSTM  & \textbf{0.0374 / 0.0677 / 0.1005}  & \textbf{0.0894 / 0.1402 / 0.1714} \\
\hline \hline
\end{tabular}
\end{table}

\begin{table*}[hbt!]
\begin{center}
\caption{Performance comparison with only 6 days parking data and domain adaptation (MelbCity to Rye)}
\label{tab:with_data}
\begin{tabular}{l||cc}
\hline \hline
Model                                & MAE (5/15/30 mins)       & RMSE (5/15/30 mins)      \\ 
\hline \hline
ConvLSTM                             & 0.0607 / 0.1091 / 0.1385 & 0.1222 / 0.1680 / 0.2003 \\
LSTM                                 & 0.0829 / 0.1035 / \textbf{0.1273} & 0.1261 / 0.1695 / 0.1998 \\
MLP+ADDA                           & 0.0845 / \textbf{0.1151} / 0.1774 & 0.1187 / \textbf{0.1616} / 0.2434\\
FADACS (ConvLSTM+ADDA)  & \textbf{0.0470} / 0.1216 / 0.1694 & \textbf{0.0813} / 0.1739 / \textbf{0.2229} \\
\hline \hline
\end{tabular}
\end{center}
\end{table*}

We conducted two sets of experiments based on the aforementioned baselines. The first experiment was the basic parking occupancy prediction experiment. In the first experiment, all models were trained and validated using data from the Rye dataset. This experiment mainly show the performance of existing method to parking prediction problem. For the transfer learning part, we applied our refined ADDA architecture on data from Melbourne and Rye to evaluate its performance. Namely, we chose the Melbourne data as the source domain and the Rye data as the target domain since the Melbourne dataset is much richer. We also trained an LSTM model and a ConvLSTM model on the source domain and tested their performance on the Rye data in this experiment. The summary of architecture is described in Table \ref{tab:acrhitecture}.

\subsection{Experimental Results}
In the first experiment, we compared some existing approaches to predict parking occupancy. We selected four classic approaches here: HA, MLP, LSTM and ConvLSTM. HA is a basic statistical method for estimating the parking occupancy based on averages of historical data. The strength of this method is that HA can detect periodical pattern in parking occupancy. However, it does not consider spatial dependency, temporal dependency and hidden trends in the data. Compared to HA, MLP can automatically explore trends of the parking occupancy even though it also does not consider spatio-temporal dependency. LSTM can predict parking occupancy by leveraging temporal dependency in historical data which is essential to time-series data prediction. However, as mentioned in the introduction, parking sensing not only relies on temporal dependency but is also relative to the spatial dependency. ConvLSTM can integrate spatial and temporal features into one simple end-to-end model, and Table \ref{tab:source} validates our assumptions, showing that ConvLSTM outperforms other classic parking prediction approaches for all prediction horizons. LSTM outperforms the second since it consider the temporal dependency but not spatial dependency. MLP performs better than HA but not as well as LSTM or ConvLSTM. This result suggests that both spatial and temporal dependency play a role in parking occupancy prediction, and temporal dependency seems more important, since the gap between the LSTM and MLP is much smaller than in MLP and other approaches.

The first experiment showed that ConvLSTM performs the best in parking sensing. Then, we conducted a few-shot transfer learning test to validate the effectiveness of our proposed transfer learning model with a few training samples from the target domain. Most machine learning techniques require thousands of examples to achieve good performance in parking prediction. The goal of few-shot learning is to achieve acceptable accuracy in parking sensing with a few training examples in the target domain. We compared our model to four classic approaches used in spatio-temporal transfer learning: LSTM with parameter transfer, ConvLSTM with parameter transfer, ADDA with MLP and our proposed architecture. The first and second models are based on a parameter transfer framework, which transfer the parameters trained in the source domain to the target domain. ADDA with MLP and our proposed architecture are GAN-based transfer learning framework. Table \ref{tab:with_data} shows that our approach performed the best in general. ConvLSTM with parameter transfer perform better than LSTM with parameter transfer, and ADDA with MLP performed the worst. This result validates our claim that both spatial and temporal dependency are important in parking occupancy prediction, and that adversarial learning is good at learning shared feature spaces. Additionally, it validates the importance of each component should be temporal dependency, spatial dependency and domain adaptation. Nevertheless, FADACS has complexity as an obvious drawback. In some cases, LSTM or MLP with ADDA perform better than the proposed approach. It is because the FADACS is a much more complicated model, which needs fine-tuning, and compared to FADACS, LSTM and MLP with ADDA have a simpler structure, which suggests that they are likely to be tuned more easily. Although our computational resources were limited and we did not conduct a heavy hyper-parameter tuning process, the performance of FADACS still outperformed other two models overall.

In summary, we conducted two experiments with Melbourne CBD parking data, Rye parking data and multiple contextual features. The experimental results show that our approach, which integrates spatial information, temporal information and domain adaptation outperforms other baselines. It also shows the importance of each component in predicting parking occupancy in a target domain by leveraging source domain historical data and contextual information. Additionally, in a real-world parking recommendation service, lower MAE or RMSE results suggest better prediction of parking availability. Service providers can recommend potential available parking slots to drivers, which could significantly save the cruising time. 


\section{Conclusion}
In this study, we used both implicit sensing and explicit sensing approaches to predict parking occupancy in two different cities. We propose a GAN-based ConvLSTM transfer learning framework to infer the parking occupancy in a new area with little historical parking data. We also qualitatively analysed the correlation between contextual information and parking occupancy with ten million-level real-world datasets. We compared our proposed model with the state-of-the-art spatio-temporal transfer learning approach, and the experimental results show that our proposed model can solve both of the significant challenges of spatial and temporal information integration and contextual information shared feature extraction. Our framework can be easily extended to other cities and other spatio-temporal sensor datasets as long as the data is graph-based and spatially correlated.

\section*{Acknowledgement}
We acknowledge the support of Mornington Peninsula Shire through the Smarter Cities and Suburbs Program (SCSP) grant. We also gratefully acknowledge the support of Doug Bradbrook from Mornington Peninsula Shire.

\bibliographystyle{IEEEtran}
\bibliography{IEEEexample}

\end{document}